# ENSEMBLE CLASSIFIER APPROACH IN BREAST CANCER DETECTION AND MALIGNANCY GRADING- A REVIEW

Deepti Ameta[1]

[1]Department of Computer Engineering, Banasthali University, Banasthali, India

*ABSTRACT*

*The diagnosed cases of Breast cancer is increasing annually and unfortunately getting converted into a high mortality rate. Cancer, at the early stages, is hard to detect because the malicious cells show similar properties (density) as shown by the non-malicious cells. The mortality ratio could have been minimized if the breast cancer could have been detected in its early stages. But the current systems have not been able to achieve a fully automatic system which is not just capable of detecting the breast cancer but also can detect the stage of it. Estimation of malignancy grading is important in diagnosing the degree of growth of malicious cells as well as in selecting a proper therapy for the patient. Therefore, a complete and efficient clinical decision support system is proposed which is capable of achieving breast cancer malignancy grading scheme very efficiently. The system is based on Image processing and machine learning domains. Classification Imbalance problem, a machine learning problem, occurs when instances of one class is much higher than the instances of the other class resulting in an inefficient classification of samples and hence a bad decision support system. Therefore EUSBoost, ensemble based classifier is proposed which is efficient and is able to outperform other classifiers as it takes the benefits of both-boosting algorithm with Random Undersampling techniques. Also comparison of EUSBoost with other techniques is shown in the paper.*

## 1. INTRODUCTION

The diagnosed cases of cancer is increasing annually and the maximum diagnosed cases are of the type Breast cancer among the middle aged women. For instance between 2009 and 2013 there was an increase of 2400 new diagnosed cases in India. This statistics is increasing annually and unfortunately getting converted into a larger death rate. This ratio could have been minimized to a certain level if the detection of the Breast cancer could have been made in the early stages. The probability of treating the cancer in its early stage is considered to be higher as compared to treating it in its more advanced stages. The signs of it may include the formation of lumps in the breast region, pain, fever, formation of pink or red patches, ejection of infectious fluid from the nipples and many more symptoms. But the problem with the cancer is that it is very much hard to detect. The treatment is most expected to be successful in the early stages.

In order to differentiate different stages, a grading system is used during the diagnosis process so that a proper treatment can be determined. It is not only being used in India but also in many other countries of the world. At present many techniques are used for detecting the malignant

cells.Different types of mammography techniques and laboratory tests are being performed for the purpose. A Fine Needle Aspiration biopsy test (FNA) is one of the techniques used to detect the malicious cells in the sample extracted from the questionable area from the breast through a very thin needle. The results help the doctors to select the appropriate treatment and therapy. Many techniques provide a grading scheme to the collected samples. Such techniques makes use of Machine Learning classifiers. The grading scheme is based on the grading scale proposed by Bloom and Richardson [1] and modified Scarff-Bloom-Richardson system [2] which involves the following:

1. *Degree of structural differentiation (SD):* It is a factor which describes the ability of the cells and tissues to get combined into groups based on some similarities. The cells are capable of forming the tubules and groups. If a single group is spotted in the stained slide under the microscope, then a case of Intermediate Malignancy case can be assured and if many groups are scattered throughout the observation then a High Malignancy Grade can be assigned to it.
2. *Frequency of hyperchromatic and mitotic figures (HMF):* In malignancy detection, spotting the behaviour of the nuclei of the cells is a significant task as the tendency of splitting the cell's nuclei can detect the stage of the cancer and how fast it is growing inside the body. Mitosis is a biological phenomenon in which the mother cell gets divided into two equal cells and hence fastens the spreading of the cancer.
3. *Plemorphism (P):* This factor differentiates the cells based on their color, shape and size, staining of the nuclei and their behaviour when they are stained with a chemical and observed under the microscope.

This grading scheme can be achieved by the applications of Machine Learning. But classification of malignancy grading suffers from the Class Imbalance Problem or imbalanced classification problem, which is a difficulty in Machine Learning process. We need to overcome this problem by using ensemble classifier approach and hence an automatic clinical decision support system for malignancy grading can be achieved which can detect the cancer of different stages very efficiently.

The main contributions of the work is as follows:

- A proposal for a clinical decision support system for malignancy grading (detection of the stage of the cancer)
- The solution to the imbalanced classification problem is proposed by combining the random undersampling with boosting and proposing Evolutionary undersampling boosting technique to make the system more efficient and stable.

## 2. THE CLASSIFICATION IMBALANCE PROBLEM

The proposed system is not just capable of detecting the cancer but can also provide a malignancy grading so that the detection of cancer in the early stage becomes easier. For this purpose there is a need to classify the samples which are taken from the questionable part of the breast, in different groups or classes so that the samples showing similar properties can be grouped into the same class. Several approaches of machine learning is used. The classification imbalance problem for two class create problems in the learning process. According to this problem, the number of instances of one class is very much greater than the number of instances of the other class [3]. One of the classes is under represented. Hence the recognition rate of the minority class gets worsen. Most of the machine learning algorithm works when the number of instances from both the classes are nearly equal. The present standard classifier learning algorithm focuses on getting a high accuracy rate and therefore the minority class is considered as a noise or disturbance and they show a biased nature towards the majority class. When addressing a two class imbalance problem, a confusion matrix can be used to store the correctly and incorrectly classified sets [4].

**Table1:** Confusion Matrix for two class problem

|  | **Positive Prediction** | **Negative Prediction** |
|---|---|---|
| **Positive class** | True Positive(TP) | False Negative (FN) |
| **Negative class** | False Positive (FP) | True Negative (TN) |

Historically, accuracy rate has been considered to evaluate the performance of the classifiers but this cannot be taken as an only parameter because accuracy rate weights the influence of the classes depending on the number of instances, and therefore classes with more instances have more influence on the results. [5]

$$\text{Accuracy} = \frac{TP+TN}{(TP+TN+FP+FN)}$$

So to converge towards the perfection we consider some more parameters

1. *Sensitivity /True positive rate (TPR)/ Recall*: It is the classification function which represents how correctly the proportion of positives are identified in the dataset i.e. sick people correctly identified as sick. It is given by the formula

$$\text{TPR} = \frac{TP}{(TP+FN)}$$

2. *Specificity /True negative rate*: It is a classification function which represents how correctly the proportion of negatives are classified in the dataset i.e. the percentage of healthy people who are correctly identified as not suffering from any diseases. It is given by the formula

$$\text{SPC} = \frac{TN}{(TN+FP)}$$

3. *Geometric Mean (GM):* The Geometric Mean considers the balance between the classification performances on the majority and minority class by considering both the parameters, sensitivity and specificity. It is given by the formula [4]

$$\text{Geometric Mean} = \sqrt{Specificity \times Sensitivity}$$

4. *ROC and AUC:* The receiver operating characteristic (ROC) curve and area under ROC curve (AUC) is an important parameter for accessing the overall classification performance. The ROC is a curve which shows the relationship between benefits (true positive rate) and costs (false positive rate) [6]. Over a wide range of operating points the ROC curve can be used to indicate whether a classifier is superior to another classifier or not. AUC provides a scalar measure of the performance of a classifier [7].The larger the AUC, the better is the performance over the data set [4]. It is given by the formula

$$\text{AUC} = \frac{1+True\ positive\ rate-False\ positive\ rate}{2}$$

5. *F-Measure:* It is the harmonic mean of recall and precision. The more F-measure value, the more is the precision and recall. It is given by the formula

$$\text{F-Measure} = \frac{2 \times Recall \times Precision}{Recall+Precision}$$

## 3. SOLUTIONS FOR THE CLASS IMBALANCE PROBLEM

To deal with class imbalance problem many techniques has been developed which can be categorized into four main categories [8]

1. *Algorithm level approaches*: These approaches deal with biasing the learning procedure towards the minority class [9]. This approach adapts a supervised classifier to strengthen the accuracy towards the minority class. It creates a new classifier or brings changes in the existing ones. This approach heavily relies on the nature of the classifier and they require the knowledge of the classifier used and the application domain as they are domain specific [10].
2. *Data level approaches:* These approaches rely on data resampling in order to reduce the effect of classification imbalance problem. They use a data pre-processing step which makes them independent of the classifier used [11]. Examples of these approaches include random undersampling and oversampling, sampling methods and feature selection strategies.
3. *Cost-sensitive learning:* The main idea is to assign a higher misclassification cost to the objects belonging to the minority class as compared to the misclassification costs assigned to majority class [10]. It requires both, data level transformations (adding costs to instances) and algorithm level modifications (modifying the learning process to accept costs) [12]. But it is very difficult for experts to assign costs to every class.
4. *Ensemble based approaches:* A classifier is introduced for each training set examples by a chosen machine learning algorithm, therefore, there will be k number of classifiers for each k variations of the training set and the result is produces by combining the output of all the classifiers [13]. AdaBoost, Bagging and RandomForest are some of the learning methods. Many reported works like SMOTEBoost, RUSBoost and DataBoost-IM have improved the classification performance [13].
5. *Hybrid approaches:* Most of the learning methods rely on combining many machine learning algorithms to achieve better results. The hybridization is used with other learning methods and is designed to alleviate the problem in sampling, feature sub-set selection and cost matrix optimization [13].

The approaches mentioned above cannot be used independently because of their disadvantages so hybridization with other approaches provides a better way of achieving good results and dealing with classification imbalance problem. Hence EUSBoost model is proposed which combines Evolutionary Undersampling with AdaBoost.M2 algorithm [14].

## 4. PROPOSED SOLUTION: EVOLUTIONARY UNDERSAMPLING BOOSTING

The combination of ensemble based classifier learning techniques has been proposed to overcome the classification imbalance problem in breast cancer malignancy grading. The most common approach is to introduce the data preprocessing step to balance the data distribution. Some of them are Bagging, Boosting and Hybrid-based ensembles depending on the ensemble based algorithm [8]. These methods are more versatile as compared to cost-sensitive ensembles because no setting of cost for different classes is required [4]. EUSBoost is a proposed solution which is based on Random Undersampling with Boosting. EUSBoost was able to outperform other methods because it uses the Evolutionary undersampling and provides diversity of the base classifiers [4].

### 4.1. Combining Boosting with Evolutionary Undersampling

Boosting algorithm [6] improves the performance of the weak classifiers by forcing the learners to learn the difficult examples. At each boosting iteration redistribution of the training data set is achieved by updating their respective weights for each sample [6]. To understand EUSBoost we will understand Boosting algorithm first as it provides a base to it. The boosting steps involved in AdaBoost can be shown by the following steps: [15]

**Steps: Adaptive Boosting algorithm (AdaBoost)** [15]

Let N be the total number of samples in the training data-set with two classes represented by y ϵ {0,1}.

1. Assign initial weights to each of the samples in the training data-set. Initially these weights should be assigned equally.
2. Now by using weighted resampling, randomly select a data-set with N samples from the original data-set. The samples with the higher weights have more probability to get selected.
3. Obtain a learner, *f(x)*, usually a predictive model or a classifier from the resampled data-set. Apply this newly obtained learner to the original sample data-set. If a wrong classification is done then the error_flag=1, otherwise error_flag=0 and compute the total sum of the weighted errors of the samples.
4. Calculate confidence index of the learner *f(x)* which depends on the weighted error. Now again update the weights of the training samples. If the samples are correctly identified then the weights remain unchanged while for misclassified samples, the weight gets increases.
5. Normalization of the weights is done and collectively their weights are considered. After T iterations we get T models using voting approach.

In Boosting, classifiers are learned serially using the whole training set in all the base classifiers. More attention is given to difficult instances so that the samples misclassified in the previous iterations can be classified correctly in the new iterations [4].

Evolutionary undersampling is an evolutionary prototype selection algorithm which is suitable to be used in class imbalance domains [4]. In this, a prototype selection is done, which is a sampling process aiming at reducing the reference set for the nearest neighbour classifier and also improving efficiency and supporting less storage necessity [16]. Chromosomes are ranked using a fitness function and in EUSBoost, the fitness function can be considered as a balancing between both the classes and the expected performance with the selected data subsets [17]. We need to promote diversity of the base classifiers so that we can correctly and efficiently classify the samples from the training data-set. Diversity refers to the fact that the base classifier should be composed in such a way that it gives different outputs hence improving the classification techniques. For this purpose, the fitness function of EUS is modified. But it is hard to promote diversity as EUS is a preprocessing algorithm. So diversity among the solutions is considered which assumes that the base classifiers which have learned from data-sets of different instances show more diversity. When we analyse the algorithm of EUSBoost, EUS embedded in AdaBoost.M2, we can find similarities with Boosting algorithm explained above as it provides a base to the new algorithm [4].

**Algorithm:** EUSBoost , EUS embedded in AdaBoost.M2 **Input:** Training set S= $\{x_i, y_i\}$, i=1….N; and $y_i \epsilon \{c_1, c_2\}$; T: Number of Iterations; I: Weak Learner

1. $D_1(i) \leftarrow \frac{1}{N}$ for i= 1,…..,N   {D is the weight distribution for the instance}
2. $w_{i,y}^1 \leftarrow D_1(i)$ for i= 1,…..,N , y≠$y_i$ {w, W, $q_t$ are weights computed from D that used along the algorithm}
3. for t=1 to T do
4. $W_i^t \leftarrow \sum_{y \neq y_i} w_{i,y}^t$

21

5. $q_t(i,y) \leftarrow \frac{w_{i,y}^t}{w_i^t}$ for $y \neq y_i$

6. $D_t(i) \leftarrow \frac{w_i^t}{\sum_{i=1}^N w_i^t}$

7. EUS is introduced, returning a new data-set ($S'$) which considers all the minority class instances and the selected ones from the majority class
$S' = EvolutionaryUndersampling(S);$

8. Now the weights for the new data-set are computed
$D'_t(k) \leftarrow \frac{w_i^t}{\sum_{x_i \epsilon S'} w_i^t}$ if $x_i \epsilon S'$
$\qquad\qquad 0 \qquad$ otherwise

9. After computing the final weights, the classifier is trained now. It has learned the training set instances. Here the modifications in the fitness function is achieved. The original data-set is preserves still those instances which are absent in the undersampled data-set carry no weight and are thrown away by the learning classifier
$h_t \leftarrow I(S, D'_t)$

10. $\epsilon_t \leftarrow \frac{1}{2}\sum_{i=1}^N D_t(i)(1 - h_t(x_i,y_i) + \sum_{i,y\neq y_i} q_t(i,y) h_t(x_i,y))$

11. $\beta_t = \frac{\epsilon_t}{1-\epsilon_t}$ {$\beta_t$ is the weight asigned to the $t^{th}$ classifier}

12. $w_{i,y}^{t+1} = w_{i,y}^t \beta_t \frac{1}{2}(1 + h_t(x_i,y_i) - h_t(x_i,y_i))$ for $i = 1, \ldots, N, y \neq y_i$

13. end for

**Output**: Boosted classifier

### 4.2. Experimental Analysis

The following figure explains how the experiment was carried out. It includes the description of the dataset used, the feature set that was used during the experiment, the experimental set up and the experimental results.

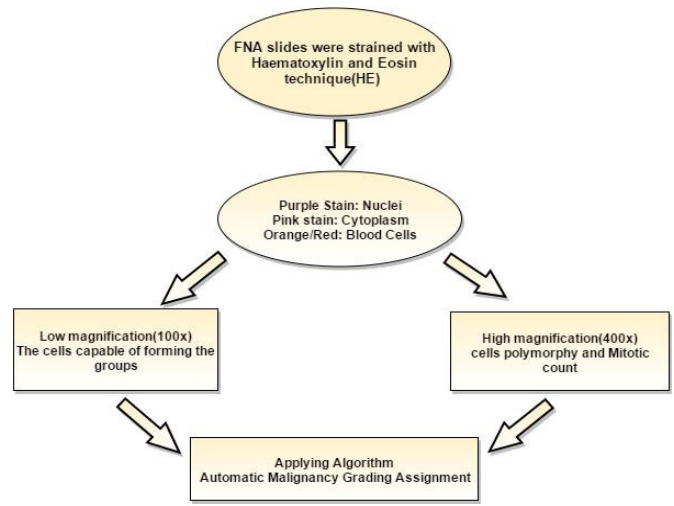

**Figure 1**: Experimental analysis of EUSBoost Technique

In this study, a dataset of FNA slides has been considered with 100 images and with a resolution of 96 dpi (dots per inch). The FNA slides were stained with Haematoxylin and Eosin technique.

International Journal of Managing Public Sector Information and Communication Technologies (IJMPICT)
Vol. 8, No. 1, March 2017After staining, the nuclei of the cells showed purple stain, cytoplasm showed pink stain and blood cells showed dark orange and red stains. Now to differentiate the classes these slides were diagnosed under two types of magnification, with low magnifying power of 100× and some were diagnosed under high magnifying power of 400× magnification. The main idea behind diagnosing them under two different magnifying power was to spot different features which were available only at different magnifying powers. Some features, like structural differentiation i.e. cells ability of forming groups, were successfully spotted at 100× magnification and some features, like mitosis and pleomorphism, were spotted under 400× magnification. The extracted features are summarized below:

**Table 2:** Extracted Features for Malignancy Grading

| The features spotted under 100× magnification | Structural differentiation, total number of groups formed, total area under the individual as well as collective groups and their distribution in the slides, nuclei and its orientation |
|---|---|
| The features spotted under 400× magnification | Mitotic count, Pleomorphism, textural features, momentum and histogram features like skew, energy and entropy. |

## 5. RESULT ANALYSIS AND CONCLUSION

After extracting the features and collecting the feature dataset, the EUSBoost algorithm was applied on a population size of 50 instances. The Geometric Mean (GM) was taken as the evaluation measure, majority was taken as the majority type with true balancing and the total number of iterations carried out were 10.As a result a Boosted classifier was gained which could classify even the ambiguous instances in the proper classes very efficiently. It helped maintaining the diversity of the base classifiers and hence provided a stability to the classification function. It also benefitted the classification scheme by providing a confidence level of the base classifiers in the weight update. The comparison of EUSBoost with other techniques is shown below:

**Table 3:** Comparison of EUSBoost with other reference methods for fuzzy c-means color segmentation

| Hypothesis | p-value (SEN) | p-value (GM) | p-value (AUC) |
|---|---|---|---|
| **EUB vs. BGG** | + (0.0031) | + (0.0042) | + (0.0072) |
| **EUB vs. BST** | + (0.0034) | + (0.0052) | + (0.0053) |
| **EUB vs. UNB** | + (0.0321) | + (0.0180) | + (0.0200) |
| **EUB vs. RBB** | + (0.0367) | + (0.0443) | + (0.0400) |
| **EUB vs. OVB** | + (0.0115) | + (0.0128) | + (0.0150) |
| **EUB vs. RUB** | + (0.0350) | + (0.0390) | + (0.0400) |

23

**Table 4:** Comparison of EUSBoost with other reference methods for grey-level quantization segmentation

| Hypothesis | p-value (SEN) | p-value (GM) | p-value (AUC) |
|---|---|---|---|
| **EUB vs. BGG** | + (0.0075) | + (0.0088) | + (0.0076) |
| **EUB vs. BST** | + (0.0095) | + (0.0112) | + (0.0110) |
| **EUB vs. UNB** | + (0.0250) | + (0.0232) | + (0.0252) |
| **EUB vs. RBB** | + (0.3484) | + (0.3555) | + (0.3512) |
| **EUB vs. OVB** | + (0.0188) | + (0.0194) | + (0.0187) |
| **EUB vs. RUB** | + (0.3488) | + (0.3575) | + (0.3517) |

**Table 5:** Overall Comparison of EUSBoost with other techniques

| *Segmentation* | *Measure* | *BGG* | *BST* | *UNB* | *RBB* | *OVB* | *RUB* | *EUB* |
|---|---|---|---|---|---|---|---|---|
| **FCM** | SEN | 71.70 | 72.8 | 88.4 | 89.98 | 84.66 | 89.8 | **92.19** |
|  | GM | 76.20 | 78.9 | 90.3 | 90.6 | 88.40 | 91.7 | **92.52** |
|  | AUC | 79 | 81 | 90.44 | 93.01 | 89 | 92.38 | **93.88** |
| **LS** | SEN | 75.90 | 74.99 | 90.24 | 92.86 | 88.05 | 91.55 | **94.61** |
|  | GM | 80.31 | 82.44 | 92.8 | 94.37 | 90.8 | 93.5 | **95.71** |
|  | AUC | 81.05 | 83.2 | 93 | 95 | 90.98 | 95.09 | **96.38** |
| **GLQ** | SEN | 70.47 | 72 | 84.99 | 90.27 | 83.92 | 90.27 | **90.27** |
|  | GM | 71.6 | 73.07 | 87.3 | 91 | 85 | 90.89 | **91.03** |
|  | AUC | 72.47 | 73.8 | 88 | 90.8 | 86.03 | 90.74 | **90.76** |

The difficulty is not only connected with the uneven distribution of the instances in the dataset but also the difficulties present in the nature of the data used. By looking at the above table it can be concluded that the standard approaches like Bagging and Boosting independently are insufficient to deliver a satisfactory classification result. Hence the classification imbalance problem can be solved with EUS method as it provides maximum efficient and satisfactory result as compared to classical models when different segmentation techniques and measures are used.

Also when we consider the learning paradigm, hybrid based approaches show poor performance for all of the measured used. Some of the experiments have proved that hybrid based approaches are competitive to other techniques but in reality, when we consider a decision support system, these techniques show the worst results. This shows that the dataset used contains some difficult samples which remains undetected for these learning classifiers. But EUS approach is able to overcome this problem as it considers the diversity among the base classifiers. Hence. EUSBoost stands superior to all other methods as it extends RUSBoost, which itself is an efficient method for solving class imbalance problem. It also applies AdaBoost.M2 algorithm for the creation of ensembles. It searches the most effective instances from the majority class and at the same time maintains the diversity among the classifiers.

Hence EUSBoost can handle the imbalanced datasets for Breast cancer Malignancy grading very efficiently. It is believed that the cancer at its early stage is very much hard to detect. But this technique is able to assign malignancy grading and helps in detection of the cancer of any stage, very importantly the cancer at the early stages.

Summarizing, it can be concluded that EUSBoost can be used to create a highly efficient and highly accurate decision support system for breast cancer malignancy grading based on ensemble classification.

## ACKNOWLEDGEMENTS

I would like to take this opportunity to express my profound gratitude and deep regards to Dr.Iti Mathur and Dr. Saurabh Mukherjee for their exemplary guidance and encouragement throughout the work. I would also like to give my sincere gratitude to all my friends and colleagues.

**Author**

**Deepti Ameta** is currently pursuing M.Tech. in Computer Science from Banasthali University, India. Her areas of interest include Machine Learning, Medical Image Processing, Soft computing and Natural Language Processing.

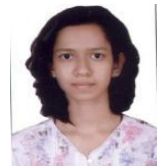